\documentclass[final]{cvpr}
\usepackage{times}
\usepackage{epsfig}
\usepackage{graphicx}
\usepackage{amsmath}
\usepackage{amssymb}
\usepackage{adjustbox}
\usepackage{booktabs}
\usepackage{multirow}

\usepackage{pifont}
\usepackage{dsfont}

\newcommand{\cmark}{\ding{51}}

\newcommand{\myparagraph}[1]{\vspace{2pt}\noindent{\bf{#1}}}

\newcommand{\ours}{CompCos}
\newcommand{\oursClosed}{\ours$^{\text{CW}}$}
\newcommand{\expandednick}{Compositional Cosine Logits}

\usepackage[pagebackref=true,breaklinks=true,colorlinks,bookmarks=false]{hyperref}

\newcommand{\myvspace}[1]{
\vspace{#1} 
}

\def\cvprPaperID{7256} 
\def\confYear{CVPR 2021}

\begin{document}

\title{Open World Compositional Zero-Shot Learning}

\author{\vspace{0em}
\setlength\tabcolsep{0.1em}
\begin{tabular}{cccc} 
Massimiliano Mancini$^{1\,*}$, & Muhammad Ferjad Naeem$^{1, 2\,*}$, & Yongqin Xian$^{3}$, Zeynep Akata$^{1,3,4}$ \tabularnewline
\end{tabular}
\\
\renewcommand{\arraystretch}{0.5}
\begin{tabular}{ccccc} 
    $^1$University of T\"{ubingen} & $^2$TU M\"{u}nchen & $^3$MPI for Informatics & $^4$MPI for Intelligent Systems 
\end{tabular}
}

\maketitle

\footnotetext[1]{First and second author contributed equally.}
\thispagestyle{empty}

\begin{abstract}
Compositional Zero-Shot learning (CZSL) requires to recognize state-object compositions unseen during training. 
In this work, instead of assuming prior knowledge about the unseen compositions, 
we operate in the open world setting, where the search space includes a large number of unseen compositions some of which might be unfeasible. In this setting, we start from the cosine similarity between visual features and compositional embeddings. After estimating the feasibility score of each composition,  we use these scores to either directly mask the output space or as a margin for the cosine similarity between visual features and compositional embeddings during training. Our experiments on two standard CZSL benchmarks show that all the methods suffer severe performance degradation when applied in the open world setting. While our simple CZSL model achieves state-of-the-art performances in the closed world scenario, our feasibility scores boost the performance of our approach in the open world setting, clearly outperforming the previous state of the art.
\end{abstract}

\section{Introduction}
The appearance of an object in the visual world is determined by its state. 
A \emph{pureed tomato} looks different from a \emph{wet tomato} despite the shared object, and a \emph{wet tomato} looks different from a  \emph{wet dog} despite the shared state. 
In Compositional Zero-Shot Learning (CZSL) \cite{misra2017redwine,nagarajan2018attributeasoperators,purushwalkam2019tmn,li2020symnet} the goal is to learn a set of states and objects while generalizing to unseen compositions.

Current benchmarks in CZSL study this problem in a closed space, assuming the knowledge of unseen compositions that might arise at test time.
For example, the widely adopted MIT states dataset \cite{isola2015mitstates} contains 28175 possible compositions (in total 115 states and 245 objects), but the test time search space is limited to 1662 compositions (1262 seen and 400 unseen), covering less than 6\% of the whole compositional space. 
This restriction on the output space is a fundamental limitation of current CZSL methods. 

\begin{figure}[t]
\centering
 \includegraphics[width=\columnwidth]{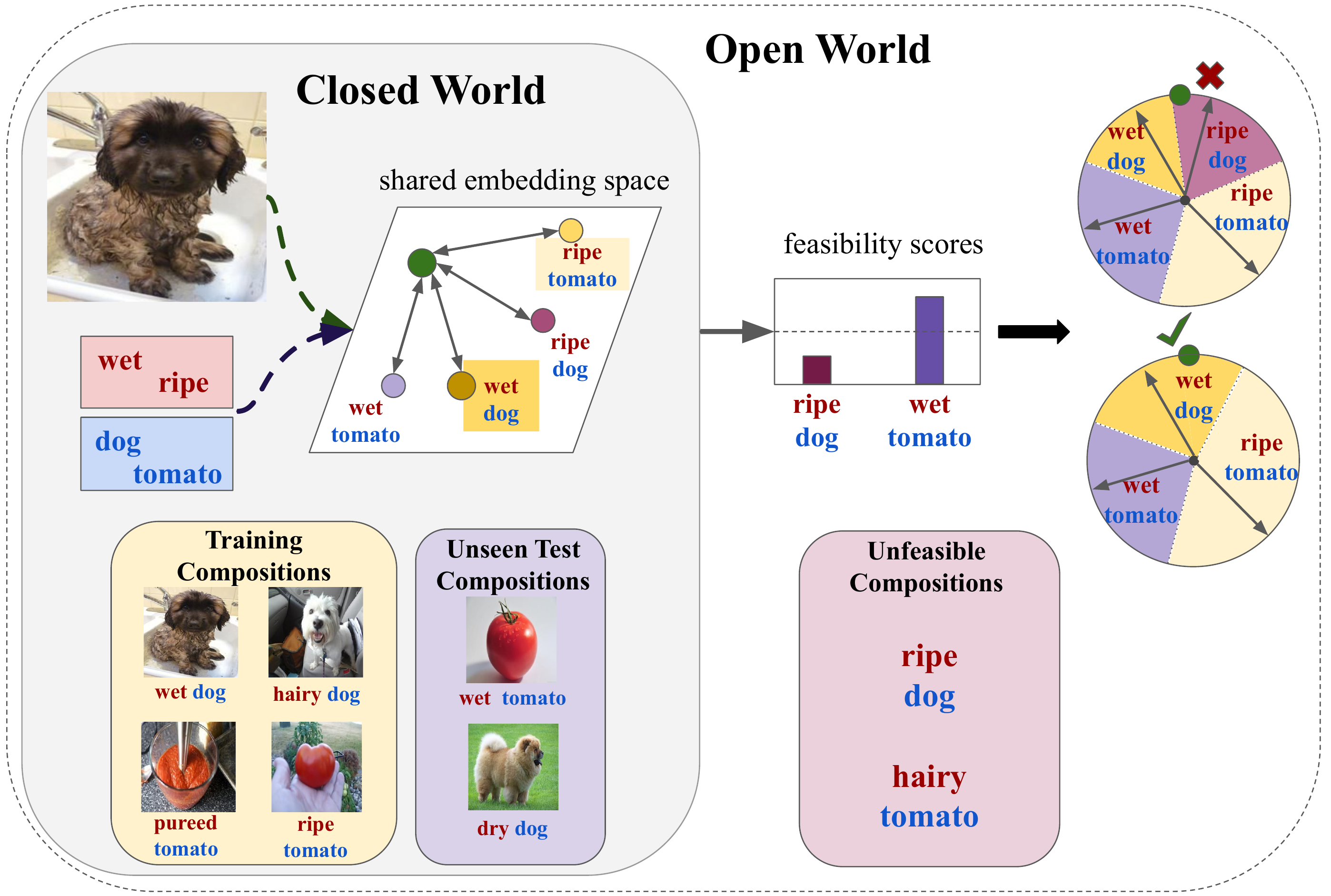}
\caption{In closed world CZSL, the search space is \textit{assumed to be known a priori}, \ie seen (yellow) and unseen (purple) compositions are available during training/test. In our open world scenario, \textit{no limit} on the search space is imposed. Hence, the model has to figure out implausible compositions (pink) and discard them.}
\myvspace{-10pt}
\label{fig:teaser}
\end{figure}

In this work, we propose the more realistic Open World CZSL (OW-CZSL) task  (see Figure \ref{fig:teaser}), where we impose \emph{no constraint} on the test time search space, causing the current state-of-the-art approaches to suffer severe performance degradation.  
To tackle this new task, we propose \emph{\expandednick} (\ours), a model where we embed both images and compositional representations into a shared embedding space and compute scores for each composition with the cosine similarity. Moreover, we treat less feasible compositions (\eg \textit{ripe dog}) as \textit{distractors} that a model needs to eliminate. For this purpose, we use similarities among primitives to assign a feasibility score to each unseen composition. We then use these scores as margins in a cross-entropy loss, showing how the feasibility scores enforce a shared embedding space where unfeasible distractors are discarded, while visual and compositional domains are aligned.  
Despite its simplicity, our model surpasses the previous state-of-the-art methods on the standard CZSL benchmark as well as in the challenging OW-CZSL task.

Our contributions are as follows: (1) A novel problem formulation, Open World Compositional Zero-Shot learning (OW-CZSL) with the most flexible search space in terms of seen/unseen compositions; (2) \ours, a novel model to solve the OW-CZSL task based on cosine logits and {a projection of} learned primitive embeddings with an integrated feasibility {estimation}  mechanism; (3) A significantly improved state-of-the-art performance on MIT states \cite{isola2015mitstates} and UT Zappos \cite{yu2014zappos,yu2017zappos} both on the existing benchmarks and the newly proposed OW-CZSL setting.

\section{Related works}
\myparagraph{Compositional Zero-Shot Learning.}
Early vision works encode compositionality with hierarchical part-based models, to learn robust and scalable object representations \cite{ommer2007compositional,ommer2009compositional,fidler2007compositional,zhu2010partcompositional,ott2011shareddpmcompositional,si2013compositional}. More recently, compositionality has been widely considered in multiple tasks such as compositional reasoning for visual question answering \cite{johnson2017clevrvqa,koushik2017vqacompositional,hudson2018vqacompositional} and modular image generation \cite{zhao2018modulargencompose,tan2019gencompositional,papadopoulos2019pizzagencompositional}. 

In this work, we focus on Compositional Zero-Shot Learning (CZSL) \cite{misra2017redwine}. Given a training set containing a set of state-object compositions, the goal of CZSL is to recognize unseen compositions of these states and objects at test time. Some approaches address this task by learning objects and states classifier in isolation and composing them to build the final recognition model. In this context, \cite{chen2014inferring} trains an SVM classifier for seen compositions and infers class weights for new compositions through a Bayesian framework. LabelEmbed \cite{misra2017redwine} learns a transformation network on top of pretrained state and object classifiers. \cite{nagarajan2018attributeasoperators} proposes to encode objects as vectors and states as linear operators that change this vector.
Similarly, \cite{li2020symnet} enforces symmetries in the representation of objects given their state transformations. Recently, \cite{purushwalkam2019tmn} proposed a modular network where states and objects are simultaneously encoded. The network blocks are then selectively activated by a gating function, taking as input an object-state composition.  

All of these works assume that the training and test-time compositions are known a priori.
We show that removing this assumption causes severe performance degradation. Furthermore, we propose the first approach for Open World CZSL. 
There are some similarities between our model and \cite{misra2017redwine} (\ie primitive representations are concatenated and projected in a shared visual-semantic space). However, our loss formulation leads to significant improvements in the Open World CZSL results. More importantly, our approach is the first to estimate the feasibility of each composition and exploits this information to isolate/remove possible distractors in the shared output space.

\myparagraph{Open World Recognition.}
In our open world setting, all the combinations of states and objects can form a valid compositional class.
This is different from an alternate definition of \textit{Open World Recognition} (OWR) \cite{bendale2015towardsowr} where the goal is to dynamically update a model trained on a subset of classes to recognize increasingly more concepts as new data arrives. 
Our definition of \textit{open world} is related to the \textit{open set} zero-shot learning (ZSL) \cite{xian2018zero} scenario in \cite{fu2016semi,fu2019vocabulary}, proposing to expand the output space to include a very large vocabulary of semantic concepts. 

Our problem formulation and approach are close in spirit to that of \cite{fu2019vocabulary} since both works consider the lack of constraints in the output space for unseen concepts as a requirement for practical (compositional) ZSL methods. 
However, there are fundamental differences between our work and \cite{fu2019vocabulary}. Since we consider the problem of CZSL, we have access to images of all primitives during training but not all their possible compositions. This implies that we can use the knowledge obtained from the visual world to model the feasibility of compositions and modifying the representations in the shared visual-compositional embedding space. We explicitly model the feasibility of each unseen composition, incorporating this knowledge into training and test.

\section{\expandednick}
\label{sec:method}
\subsection{(OW)-CZSL Task Definition}
\label{sec:problem-definition}
Compositional zero-shot learning (CZSL) aims to predict a composition of multiple semantic concepts in images. Let us denote with $\mathcal{S}$ the set of possible states, with $\mathcal{O}$ the set of possible objects, and with $\mathcal{C}=\mathcal{S}\times\mathcal{O}$ the set of all their possible compositions. 
$\mathcal{T}=\{(x_i,c_i)\}_{i=1}^N$ is a training set where $x_i\in\mathcal{X}$ is a sample in the input (image) space $\mathcal{X}$ and $c_i\in\mathcal{C}^s$ is a composition in the subset $\mathcal{C}^s\subset\mathcal{C}$. 
$\mathcal{T}$ is used to train a model $f:\mathcal{X}\rightarrow \mathcal{C}^t$ predicting combinations in a space $\mathcal{C}^t\subseteq\mathcal{C}$ where $\mathcal{C}^t$ may include compositions that are not present in $\mathcal{C}^s$~(i.e. $\exists c \in \mathcal{C}^t \land c\notin \mathcal{C}^s$).

The CZSL task entails different challenges depending on the extent of the target set $\mathcal{C}^t$. If $\mathcal{C}^t$ is a subset of $\mathcal{C}$ and $\mathcal{C}^t\cap \mathcal{C}^s \equiv \emptyset$, the task definition is of \cite{misra2017redwine}, where the model needs to predict only unseen compositions at test time. In case $\mathcal{C}^s\subset\mathcal{C}^t$ we are in the generalized CZSL scenario, and the output space of the model contains both seen and unseen compositions. Similarly to the standard generalized zero-shot learning \cite{xian2018zero}, this scenario is more challenging due to the natural prediction bias of the model in $\mathcal{C}^s$, seen during training. Most recents works on CZSL consider the generalized scenario \cite{purushwalkam2019tmn,li2020symnet}, and the set of unseen compositions in $\mathcal{C}^t$ is assumed to be known a priori, with $\mathcal{C}^t\subset\mathcal{C}$. 

\begin{figure*}[t]
 \includegraphics[width=\linewidth]{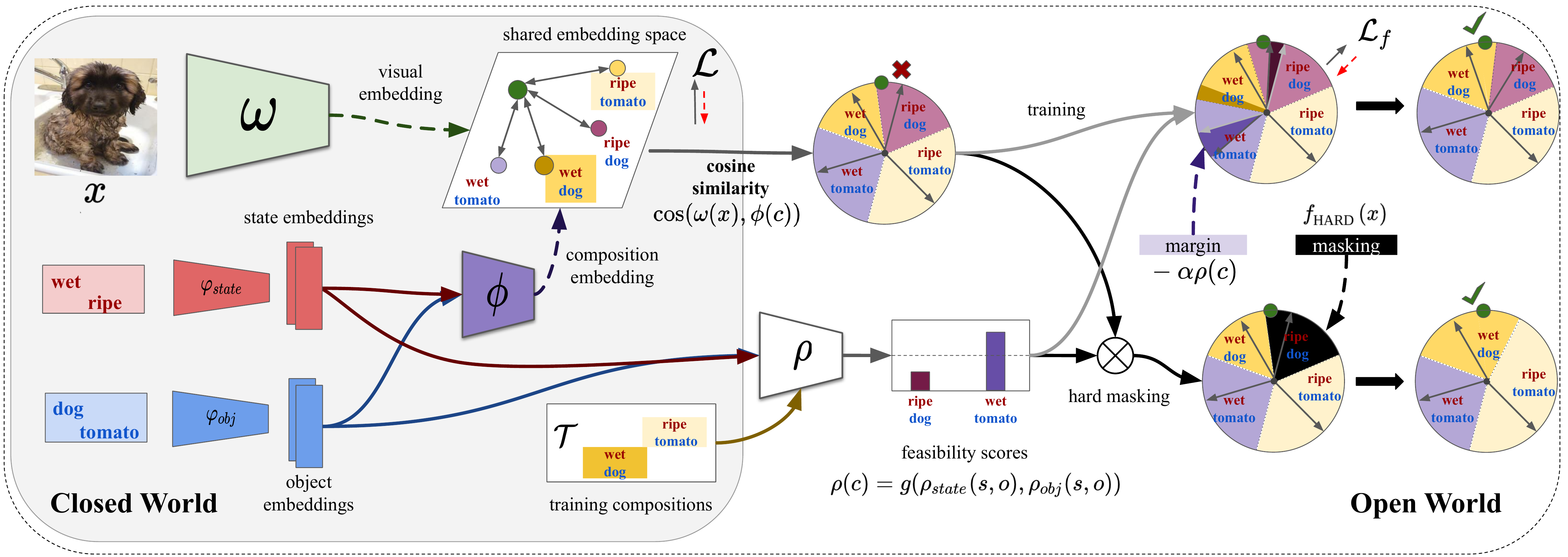}
\caption{\textbf{\expandednick\ (\ours).} Our approach embeds an image (top) and state-object compositions (bottom) into a shared semantic space defined by the cosine-similarity between image features and composition embeddings. In the open world model, we estimate a feasibility score for each of the unseen compositions, using the relation between states, objects, and the training compositions. The feasibility scores are injected to the model either by removing less feasible compositions (\eg \textit{ripe dog}) from the output space (bottom, black slice) or by adding a bias to the cosine similarities computed during training (top, purple slices).
}
\myvspace{-10pt}
\label{fig:method}
\end{figure*}

In this work, we take a step further, analyzing the case where the output space of the model is the whole set of possible compositions $\mathcal{C}^t\equiv\mathcal{C}$, i.e. \textit{Open World Compositional Zero-shot Learning} (OW-CZSL). Note that this task presents the same challenges of the generalized case while being far more difficult since i) $|\mathcal{C}^t|\gg|\mathcal{C}^s|$, thus it is hard to generalize from a small set of seen to a very large set of unseen compositions; and ii) there are a large number of \textit{distractor} compositions in $\mathcal{C}^t$, \ie compositions predicted by the model but not present in the actual test set that can be close to other unseen compositions, hampering their discriminability. We highlight that, despite being similar to Open Set Zero-shot Learning \cite{fu2019vocabulary}, we do not only consider objects but also states. Therefore, this knowledge can be exploited to identify unfeasible distractor compositions (\eg \textit{rusty pie}) and isolate them. 
Figure \ref{fig:method} shows an overview of our approach for both closed and open world scenarios.

\subsection{{\ours}: A Closed World Model}
\label{sec:compcos-closed}
In this section, we focus on the closed world setting, where $\mathcal{C}^s\subset\mathcal{C}^t\subset\mathcal{C}$. Since in this scenario $|\mathcal{C}^t|\ll|\mathcal{C}|$ and the number of unseen compositions is usually lower than the number of seen ones, 
this problem presents several challenges. 
In particular, while learning a mapping from the visual to the compositional space, the model needs to avoid being overly biased toward seen class predictions. Inspired by incremental learning \cite{hou2019learning} and generalized few-shot learning \cite{gidaris2018dynamic}, we reduce this problem by replacing the logits of the classification layer with cosine similarities between the image features and the composition embeddings in the shared embedding space: 
\begin{equation}
\label{eq:prediction}
f(x) = \arg\max_{c\in\mathcal{C}^t} cos(\omega(x),\phi(c))
\end{equation}
where $\omega:\mathcal{X}\rightarrow\mathcal{Z}$ is the mapping from the image space to the shared embedding space $\mathcal{Z}\in \mathbb{R}^d$ and $\phi:\mathcal{C}\rightarrow\mathcal{Z}$ embeds a composition to the same space. $cos(y,z)=\frac{y^\intercal z}{||y||\,||z||}$ is the cosine similarity among the two embeddings. 
As sketched Figure \ref{fig:method} (left), the visual embedding $\omega$ (green block) maps an image to the shared embedding space, while states (red) and objects (blue) embeddings, are embedded to the shared space by $\phi$ (purple block). 

\myparagraph{Visual embedding.} We use a standard deep neural network, e.g. ResNet-18~\cite{he2016deep}, with an additional embedding function $\omega$ mapping, the feature extracted from the backbone to $\mathcal{Z}$. The embedding function is a simple 2-layer MLP, where dropout \cite{srivastava2014dropout}, LayerNorm \cite{ba2016layer} and a ReLU non-linearity \cite{nair2010rectified} are applied after the first layer. During training, we freeze the main backbone, updating only the final MLP. 

\myparagraph{Composition embedding.} The function $\varphi:\mathcal{S}\cup \mathcal{O}\rightarrow \mathbb{R}^d$ maps the primitives, \ie objects and states, into their corresponding embedding vectors. The embedding of a given composition $c=(s,o)$ is a simple linear projection of the embeddings of its primitives:
\[
\phi(c) = [\varphi(s)\;\varphi(o)]^\top W
\]
with $W\in\mathbb{R}^{2d\times d}$, 
where we consider $\mathcal{Z}\in\mathbb{R}^d$ for simplicity. 
We chose a linear embedding function since we found it works well in practice. Moreover, it applies a strong constraint to the compositional space, making 
the embedding less prone to overfitting and more suitable for generalizing in a scenario where we might have $|\mathcal{C}^t|\gg|\mathcal{C}^s|$. During training, we update both the embedding matrix $W$ and the atomic embeddings of $\varphi$, after initializing the latter with word embeddings. 
\myparagraph{Objective function.} We define a cross-entropy loss on top of the cosine logits: 
\begin{equation}
\label{eq:objective}
    \mathcal{L} = -\frac{1}{|\mathcal{T}|}\sum_{(x,c)\in\mathcal{T}} \log\frac{e^{\frac{1}{T}\cdot p(x,c)}}{\sum_{y\in\mathcal{C}^s}e^{\frac{1}{T}\cdot p(x,y)}}
\end{equation}
where $T$ is a temperature value that balances the probabilities for the cross-entropy loss \cite{zhang2019adacos} and $p(x,c)=\cos(\phi(x),\omega(c))$. In the following we discuss how to extend our \expandednick\ (\ours) model from the closed to the more challenging open world scenario. 

\subsection{\ours: from Closed to Open World}
\label{sec:compcos-open}

Although \ours\ is an effective CZSL algorithm in the standard closed world scenario, performing well on the OW-CZSL requires tackling different challenges, such as avoiding distractors. We consider distractors as less-likely concepts, e.g. \textit{pureed dog}, \textit{hairy tomato}, and argue that the similarity among objects and states can be used as a proxy to estimate the feasibility of each composition. The least feasible compositions can then be successfully isolated, improving the representations of the feasible ones.

\myparagraph{Estimating Compositional Feasibility.}
Let us consider two objects, namely \textit{cat} and \textit{dog}. We know, from our training set, that \textit{cats} can be \textit{small} and \textit{dogs} can be \textit{wet} since we have at least one image for each of these compositions. However, the training set may not contain images for \textit{wet cats} and \textit{small dogs}, which we know are feasible in reality. We conjecture, that similar objects share similar states while dissimilar ones do not. Hence, it is safe to assume that the states of \textit{cats} can be transferred to \textit{dogs} and vice-versa. 

With this idea in mind, given a composition $c=(s,o)$ we define its feasibility score with respect to the object $o$ as:
\begin{equation}
\label{eq:obj}
\rho_{obj}(s,o)= \max_{\hat{o}\in\mathcal{O}^s} cos(\varphi(o),\varphi(\hat{o}))
\end{equation}
with $\mathcal{O}^s$ being the set of objects associated with state $s$ in the training set $\mathcal{C}^s$, \ie $\mathcal{O}^s =\{o | (s,o) \in \mathcal{C}^s\}$. Note, that the score is computed as the cosine similarity between the object embedding and the most similar other object with the target state, thus the score is bounded in $[-1,1]$. Training compositions get assigned the score of 1.
Similarly, we define the score with respect to the state $s$ as: 
\begin{equation}
\label{eq:attr}
\rho_{state}(s,o)= \max_{\hat{s}\in\mathcal{S}^o} cos(\varphi(s),\varphi(\hat{s}))
\end{equation}
with $\mathcal{S}^o$ being the set of states associated with the object $o$ in the training set $\mathcal{C}^s$, \ie $\mathcal{S}^o =\{s | (s,o) \in \mathcal{C}^s\}$. 

The feasibility score for a composition $c=(s,o)$ is then:
\begin{equation}
\label{eq:final-score}
\rho(c)=\rho(s,o)= g(\rho_{state}(s,o), \rho_{obj}(s,o))
\end{equation}
where $g$ is a mixing function, \eg max operation ($g(x,y)=\max(x,y)$) or the average ($g(x,y)=(x+y)/2$), keeping the feasibility score bounded in $[-1,1]$. Note that, while we focus on extracting feasibility from the visual information, 
external knowledge (\eg knowledge bases \cite{liu2004conceptnet}, language models \cite{wang2019language}) can be complementary resources.

\myparagraph{Exploiting Compositional Feasibility.}
A first simple strategy is applying a threshold on the feasibility scores, considering all compositions above the threshold as valid and others as distractors (e.g. \textit{ripe dog}, as shown in the black pie slice of Figure \ref{fig:method}):
\begin{equation}
\label{eq:score-hard}
f_{\text{HARD}}(x) = \underset{c\in\mathcal{C}^t, \rho(c)>\tau}{\arg\max} \,\, cos(\omega(x),\phi(c))
\myvspace{-3pt}
\end{equation}

where 
$\tau$ is the threshold, tuned on a validation set. While this strategy is effective, it might be too restrictive in practice. For instance, \textit{tomatoes} and \textit{dogs} being far in the embedding space does not mean that a state for \textit{dog}, e.g. \textit{wet}, cannot be applied to a \textit{tomato}. Therefore, considering the feasibility scores as the golden standard may lead to excluding valid compositions. 
To sidestep this issue, we propose to inject the feasibility scores directly into the training procedure. We argue that doing so 
can enforce separation between most and least feasible unseen compositions in the shared embedding space. 

To inject the feasibility scores $\rho(c)$  directly within our objective function, we can define:
\begin{equation}
\label{eq:objective-feasibility}
    \mathcal{L}_{\text{f}} = -\frac{1}{|\mathcal{T}|}\sum_{(x,c)\in\mathcal{T}} \log\frac{e^{\frac{1}{T}\cdot p^f(x,c)}}{\sum_{y\in\mathcal{C}}  e^{\frac{1}{T}\cdot p^f(x,y)}}
    \end{equation}
with:
\begin{equation}
 \label{eq:margin-scores}
    p^f(x,c)= \begin{cases}
    \cos(\omega(x),\phi(c)) & \text{if}\,\, c\in\mathcal{C}^s\\
    \cos(\omega(x),\phi(c)) - \alpha\rho(c) &\text{otherwise}
    \end{cases}
\end{equation}
where $\rho(c)$ are used as margins for the cosine similarities, and $\alpha>0$ is a scalar factor.
With Eq.~\eqref{eq:objective-feasibility} we include the full compositional space while training with the seen compositions data to raise awareness of the margins between seen and unseen compositions directly during training. 

Note that, since $\rho(c_i)\neq\rho(c_j)$ if $c_i\neq c_j$ and $c_i,c_j \notin \mathcal{C}^s$, we have a different margin, i.e. $-\alpha\rho(c)$, for each unseen composition $c$. 
This is because most feasible compositions should be closer to the seen ones (to which the visual embedding network is biased) than less feasible ones. 
By doing that, we force the network to push the representation of less feasible compositions away from the representation of compositions in $\mathcal{C}^s$ in $\mathcal{Z}$. On the other hand, the less penalized feasible compositions benefit from the updates produced for seen training compositions containing the same primitives.  
More feasible compositions will then be more likely to be predicted by the model, even without being present in the training set. As an example (Figure \ref{fig:method}, top part), the unfeasible composition \textit{ripe dog} is more penalized than the feasible \textit{wet tomato} during training, with the outcome that the optimization procedure does not force the model to reduce the region of \textit{wet tomato}, while reducing the one of \textit{ripe dog} (top-right pie).

We highlight that in this stage we do not explicitly bound the revised score $p^f(c)$ to $[-1,1]$. Instead, we let the network implicitly adjust the cosine similarity scores during training. We also found it beneficial to linearly increase $\alpha$ till a maximum value as the training progresses, rather than keeping it fixed. This permits us to gradually introduce the feasibility margins within our objective while exploiting improved primitive embeddings to compute them. 

\section{Experiments}
\setlength{\tabcolsep}{2pt}
\renewcommand{\arraystretch}{1.2}
\begin{table*}[t]
    \centering
    \resizebox{\linewidth}{!}{\begin{tabular}{l| c c c c c c | c c c c c c | c c c c c c | c c c c c c}
    \multirow{3}{*}{\textbf{Method}} 
    & \multicolumn{12}{c}{\textbf{Closed World}}& \multicolumn{12}{|c}{\textbf{Open World}}\\
    & \multicolumn{6}{c}{\textbf{MIT states}}& \multicolumn{6}{c}{\textbf{UT Zappos}}& \multicolumn{6}{|c}{\textbf{MIT states}}& \multicolumn{6}{c}{\textbf{UT Zappos}}\\
                                                &Sta. &Obj.    &S   & U    & HM    & auc &Sta. &Obj.    &S   & U    & HM    & auc &Sta. &Obj.    &S   & U    & HM    & auc &Sta. &Obj.    &S   & U    & HM    & auc \\\hline
     AoP\cite{nagarajan2018attributeasoperators}&21.1   & 23.6      &14.3   &17.4       &9.9    &1.6 &38.9   & 69.9      &{59.8}   &54.2       &40.8    &25.9
     &15.4   & 20.0      &16.6   &5.7       &4.7    &0.7 & 25.7  &   61.3    &  50.9 &  34.2     & 29.4   & 13.7\\
     LE+\cite{misra2017redwine}                 &23.5   & 26.3      &15.0   &20.1       &10.7   &2.0 &41.2   & 69.3      &53.0   &{61.9}       &41.0   &25.7&
     10.9   & 21.5      &14.2   &2.5       &2.7   &0.3  & 
     \textbf{38.1}  &  68.2     & \textbf{60.4}  &   36.5    & 30.5  & 16.3            \\
     TMN\cite{purushwalkam2019tmn}              &23.3   & 26.5      &20.2   &20.1       &13.0   &2.9 &40.8   & 69.5      &58.7   &60.0       &\textbf{45.0 }  &\textbf{29.3}
     &  6.1   &   15.9   &  12.6        &  0.9     &   1.2        &   0.1  &  14.6    &         61.5 &     55.9  &18.1           & 21.7       &  8.4\\
     SymNet\cite{li2020symnet}                  &26.3   & 28.3      &24.2   &\textbf{25.2}       &{16.1}   &3.0 &41.3   & {68.6}      &49.8   &{57.4}       &{40.4}   &23.4
     &17.0   & 26.3      &21.4   &7.0       &5.8   &0.8              &33.2   &70.0       &53.3   &44.6       &34.5   &18.5\\\hline
     \textbf{\ours}                             &\textbf{27.9}   & \textbf{31.8}      &\textbf{25.3}   &24.6       &\textbf{16.4}   &\textbf{4.5} &\textbf{{44.7}}   & \textbf{73.5}      &\textbf{59.8}   &\textbf{62.5}       &{43.1}   &{28.7}
     &\textbf{18.8}   & \textbf{27.7}      &\textbf{25.4}   &\textbf{10.0}       &\textbf{8.9}   &\textbf{1.6} & 35.1 & \textbf{72.4} & 59.3 & \textbf{46.8} & \textbf{36.9} & \textbf{21.3}\\
    \end{tabular}}
    \vspace{1pt}
    \caption{Closed and Open World CZSL results on MIT states and UT Zappos. We measure states (Sta.) and objects (Obj.) accuracy on the primitives, best seen (S) and unseen accuracy (U), best harmonic mean (HM), and area under the curve (auc) on the compositions.}
    \myvspace{-8pt}
    \label{tab:sota}
\end{table*}

 \myvspace{-4pt}
\myparagraph{Datasets.}
We experiment with two standard CZSL benchmark datasets. 
MIT states \cite{isola2015mitstates} contains 53K images of 245 objects in 115 possible states. We adopt the standard split from~\cite{purushwalkam2019tmn}. For the closed world experiments, the output space contains 1262 seen and 300/400 unseen (validation/test) compositions. For the open world scenario, we consider all possible 28175 compositions as present in the search space. Note, that 26114 out of 28175 ($\sim$93\%) are not present in any splits of the dataset but are included in our open world setting.

UT Zappos \cite{yu2014zappos,yu2017zappos} contains 12 different shoe types (objects) and 16 different materials (states). In the closed world setting, the output space is constrained to the 83 seen and to additional 15 and 18 unseen compositions for validation and test respectively. Although 76 out of 192 possible compositions ($\sim$40\%) are not in any of the splits of the dataset, we consider them in our open world setting.
 
 \myvspace{-1pt}
\myparagraph{Metrics.} 
For the primitives, we report the object and state classification accuracies. Since we focus on the generalized scenario and the model has an inherent bias for seen compositions, we follow the evaluation protocol from \cite{purushwalkam2019tmn}. We consider the performance of the model with different bias factors for the unseen compositions, reporting the results as best accuracy on only images of seen compositions (\textit{best seen}), best accuracy on only unseen compositions (\textit{best unseen}), best harmonic mean (\textit{best HM}) and Area Under the Curve (AUC) for seen and unseen accuracies at bias values.

 \myvspace{-1pt}
\myparagraph{Benchmark and Implementation Details.}
As in \cite{purushwalkam2019tmn,li2020symnet} our image features are extracted from a ResNet18 pretrained on ImageNet~\cite{deng2009imagenet} and we learn the visual embedding module $\omega$ on top of these features. We initialize the embedding function $\varphi$ with 300-dimensional {word2vec}~\cite{mikolov2013distributed} embeddings for UT Zappos and with 600-dimensional {word2vec+fastext}~\cite{bojanowski2017enriching} embeddings for MIT states, following \cite{xian2019semantic}, keeping the same dimensions for the shared embedding space $\mathcal{Z}$. We train both $\omega$ ($\varphi$ and $W$) and $\phi$ using Adam~\cite{kingma2014adam} optimizer with a learning rate and a weight decay set to $5\cdot10^{-5}$. The margin factor $\alpha$ and the temperature $T$ are set to $0.4$ and $0.05$ respectively for MIT states and $1.0$ and $0.02$ for UT Zappos. We linearly increase $\alpha$ from 0 to the previous values during training,  reaching the values after 15 epochs. We consider the mixing function $g$ as the average to merge state and object feasibility scores and $f_\text{HARD}$ as predictor for OW-CZSL, unless otherwise stated. 

We compare with four state-of-the-art methods, Attribute as Operators (AOP) \cite{nagarajan2018attributeasoperators}, considering objects as vectors and states as matrices modifying them \cite{nagarajan2018attributeasoperators}; LabelEmbed+ (LE+) \cite{misra2017redwine,nagarajan2018attributeasoperators} training a classifier merging state and object embeddings with an MLP; Task-Modular Neural Networks (TMN) \cite{purushwalkam2019tmn}, modifying the classifier through a gating function receiving as input the queried stat-object composition; and SymNet \cite{li2020symnet}, learning object embeddings showing symmetry under different state-based transformations. 
We train each model with their default hyperparameters, reporting the closed and open world results of the models with the best AUC on the validation set.

\subsection{Comparing with the State of the Art}

We compare \ours\ and the state of the art in the standard closed world and the proposed open world setting.

\myparagraph{Closed World Results.} The results analyzing the performance of \ours\ in the standard closed world scenario are reported on the left side of Table \ref{tab:sota} for the MIT states and UT Zappos test sets. Although being a substantially simple approach, our \ours\ achieves remarkable results. On MIT states our model either outperforms or is comparable to all competitors in all metrics. In particular, while obtaining a comparable best harmonic mean with SymNet, it achieves 4.5 AUC, which is a significant improvement over 3.0 from SymNet. This highlights how our model is more robust to the bias on unseen test compositions. Compared to the closest method to ours, LabelEmbed+ (LE+), \ours\ shows clear advantages for all metrics (from 2 to 4.5 AUC, and from 10.7\% to 16.4\% on the best harmonic mean), underlying the impact of our embedding functions and the cross-entropy loss on cosine logits.

On UT Zappos, our model is superior to almost all methods (except TMN in two cases). It is particularly interesting how \ours\ surpasses AoP, LE+, and SymNet with more than $2\%$ on best harmonic mean and at least by $2.8$ in AUC. In comparison with TMN, while achieving a lower best harmonic mean (-1.9\%) and a slightly lower AUC (-0.6), it achieves the best unseen accuracy (+2.5\%) and improves of $4\%$ the accuracy in recognizing each primitive in isolation. These results show that \ours\ is less sensitive to the value of the bias applied to the unseen compositions in the generalized scenario, thanks to the use of cosine similarity as prediction score. We would like to highlight that our model uses a magnitude lower trainable parameters, e.g. 0.8M vs 2.3M for TMN, to achieve these results.

\myparagraph{Open World Results.} The results on the challenging OW-CZSL setting are 
reported on the right side of Table \ref{tab:sota}. As expected, the first clear outcome is the severe decrease in performance of every method. In fact, the OW-CZSL performances (\eg best unseen, best HM, and AUC) are less than half of CZSL performances in MIT states. The largest decrease in performance is on the best unseen metric, due to the presence of a large number of distractors. As an example, LE+ goes from 20.1\% to 2.5\% of best unseen accuracy and even the previous state of the art, SymNet, loses 18.2\%, confirming that the open world scenario is significantly more challenging than the closed world setting.

In the OW-CZSL setting, our model (\ours) is more robust to the distractors, due to the injected feasibility-based margins which shape the shared embedding space during training. 
This is clear in MIT states, where \ours\ outperforms the state of the art for all metrics. Remarkably, it obtains double the AUC of the best competitor, SymNet, going from 0.8 to 1.6 with a 3.1\% improvement on the best HM and $3.0\%$ on best unseen accuracy. 

In UT Zappos the performance gap with the other approaches is more nuanced. This is because the vast majority of compositions in UTZappos are feasible, thus it is hard to see a clear gain {from injecting} the feasibility scores into the training procedure. Nevertheless, \ours\ improves by 2.8 in AUC and 2.4 in best unseen accuracy over SymNet, showing the highest results according to all compositional metrics but the seen accuracy, where it performs comparably to LE+. This is expected, since the output space of all previous works is limited to seen classes during training, thus the models are {discriminative for seen compositions}.

 \setlength{\tabcolsep}{3pt}
\renewcommand{\arraystretch}{1.2}
\begin{table}[t]
    \centering
    \resizebox{\linewidth}{!}{\begin{tabular}{l | l| c c c c}
    \hline
    \multicolumn{2}{l|}{\textbf{Effect of the Margins}}   & Seen & Unseen    & HM    & AUC\\\hline
        \oursClosed  & & 28.0 &6.0 &7.0 &1.2\\
        \ours & $\alpha=0$& 25.4 &10.0 &9.7 &1.7\\
         &+ $\alpha>0$ & 27.0 & 10.9& 10.5  &2.0 \\
         &+ warmup $\alpha$  & 27.1&{11.0}&10.8& 2.1\\
         \hline
       \multicolumn{2}{l|}{\textbf{Effect of Primitives}}   & Seen & Unseen    & HM    & AUC\\\hline
        \multirow{4}{*}{\ours} &$\rho_{state}$ & 26.6 & 10.2& 10.2&1.9\\
        &  $\rho_{obj}$  &27.2& 10.0 &9.9  &1.9\\
        &$\max(\rho_{state},\rho_{obj})$ &27.2&10.1&10.1& 2.0  \\
        &$(\rho_{state}+\rho_{obj})/2$   &27.1&{11.0}&10.8& 2.1 \\
        
    \end{tabular}}
    \vspace{1pt}
    \caption{Results on MIT states validation set for different ways of applying the margins (top) and different ways of computing the feasibility scores (bottom) for \ours\ with $f$ as predictor.}
    \myvspace{-16pt}
    \label{tab:ablation-feasibility}
\end{table}

\subsection{Ablation studies}
We investigate the impact of the feasibility-based margins, how we obtain them (without $f_{\text{HARD}}$), and the benefits of limiting the output space during inference using $f_{\text{HARD}}$. We perform our analyses on MIT states' validation set. Note that, in the tables, \oursClosed\ is the closed world \ours\ model, as described in Sec.~\ref{sec:compcos-closed}.
 
 \myvspace{-1pt}
\myparagraph{Importance of the feasibility-based margins.}  
We check the impact of including all compositions in the objective function (without any margin) and of including the feasibility margin but without any warmup strategy for $\alpha$. 

As the results in Table \ref{tab:ablation-feasibility} (Top) shows, including all unseen compositions in the cross-entropy loss without any margin (\ie $\alpha=0$) increases the best unseen accuracy by 4\% and the AUC by 0.5. This is a consequence of the training procedure: since we have no positive examples for unseen compositions, including unseen compositions during training makes the network push their representation far from seen ones in the shared embedding space. This strategy regularizes the model in presence of a large number of unseen compositions in the output space. Note, that this problem is peculiar in the open world scenario since in the closed world the number of seen compositions is usually larger than the unseen ones. The \ours\ ($\alpha=0$) model performs worse than \oursClosed~ on seen compositions, as the loss treats all unseen compositions equally.

Results increase if we include the feasibility scores during training (\ie $\alpha>0$). The AUC goes from 1.7 to 2.0, with consistent improvements over the best seen and unseen accuracy. This is a direct consequence of using the feasibility to separate the unseen compositions from the unlikely ones. In particular, this brings a large improvement on Seen and moderate improvements on both Unseen and HM.

Finally, linearly increasing $\alpha$ (\ie warmup $\alpha$) further improves the harmonic mean due to both the i) improved margins that \ours\ estimates from the updated primitive embeddings and ii) the gradual inclusion of these margins in the objective.
This strategy {improves} the bias between seen and unseen classes (as for the {better} on harmonic mean) while slightly {enhancing} the discriminability on seen and unseen compositions in isolation.

\myparagraph{Effect of Primitives} 
We can either use objects as in Eq.~\eqref{eq:obj}, states as in Eq.~\eqref{eq:attr}) or both  as in Eq.~\eqref{eq:final-score} to estimate the feasibility score for each unseen composition. Here we consider all these choices, showing their impact on the results in Table \ref{tab:ablation-feasibility} (Bottom),  with $f$ as predictor. 

The results show that computing feasibility on the primitives alone is already beneficial (achieving an AUC of 1.9) since the dominant states like \textit{caramelized} and objects like \textit{dog} provide enough information to transfer knowledge. 
In particular, computing the scores starting from state information ($\rho_{\text{state}}$) brings good best unseen and HM results while under-performing on the best seen accuracy. On the other hand, using similarities among objects ($\rho_{\text{obj}}$) performs well on the seen classes while achieving slightly lower performances on unseen ones and HM. 

Nevertheless introducing both states and objects give the best result at AUC of 2.1 as it combines the best of both. Merging objects and states scores through their maximum ($\rho_{\text{max}}$) maintains the higher seen accuracy of the object-based scores, with a trade-off between the two on unseen compositions. However, merging objects and states scores through their average brings to the best performance overall, with a significant improvement on unseen compositions (almost 1\%) as well as the harmonic mean. 
We ascribe this behavior to the fact that, with the average, the model is less-prone to assign either too low or too high feasibility scores for the unseen compositions, smoothing their scores.
As a consequence, more meaningful margins are used in Eq.~\eqref{eq:objective-feasibility} and thus the network achieves a better trade-off between discrimination capability on the seen compositions and better separating them from unseen compositions (and distractors) in the shared embedding space. 

\setlength{\tabcolsep}{3pt}
\renewcommand{\arraystretch}{1.2}
\begin{table}[t]
    \centering
    \begin{tabular}{l| c | c c c c}
       & Mask   &Seen   & Unseen    & HM    & AUC\\\hline
        \multirow{2}{*}{LE+} &&\multirow{2}{*}{14.8}& 3.1&3.2 &0.3\\
        &\cmark&& 5.0& 4.6&0.5\\\hline
                  
        \multirow{2}{*}{TMN}  &&\multirow{2}{*}{15.9}&1.3 &1.7 &0.1\\
         &\cmark&& 4.1&4.1 &0.4\\\hline
                  
         \multirow{2}{*}{SymNet}  &&\multirow{2}{*}{23.6}& 7.9&7.6 &1.2\\
         &\cmark&& 7.9& 7.7&1.2\\\hline
                  
          \multirow{2}{*}{\oursClosed}  &  &  \multirow{2}{*}{28.0} &6.0 &7.0 &1.2\\
          & \cmark  &  &8.1&8.7& 1.6 \\
          \hline
                  
         \multirow{2}{*}{\ours}  &&\multirow{2}{*}{27.1}&{11.0}&10.8& {2.1}\\
          &  \cmark & &11.2&11.0&2.2  \\
    \end{tabular}
    \vspace{1pt}
    \caption{Results on MIT states validation set for applying our feasibility-based binary masks ($f_{\text{HARD}}$) on different models.}
    \myvspace{-15pt}
    \label{tab:ablation-inference}
\end{table}

\myparagraph{Effect of Masking.} We consider mainly two ways of using the feasibility scores: during training as margins and/or during inference as masks on the predictions. We analyze the impact of applying the mask during inference, \ie using as prediction function Eq.~\eqref{eq:score-hard} in place of Eq.~\eqref{eq:prediction}, with the threshold $\tau$ computed empirically. 
We run this study on our closed world model \oursClosed, our full model \ours\ and three CZSL baselines, LE+, TMN, and SymNet. Note that, since seen compositions are not masked, best seen performances are shared across a single model.

As shown in Table \ref{tab:ablation-inference}, if we apply binary masks on top of \oursClosed\, the AUC increases by 0.5, best harmonic mean by 1.7\%, and best unseen accuracy by 2.1\%. This is because our masks filter out the less feasible compositions, rather than just restricting the output space. 
At the same time, the improvements are not as pronounced for the full \ours\ model. Indeed, including the feasibility scores as margins during training makes the model already robust to the distractors. The hard masking still provides a slight benefit over all compositional metrics, with an 11\% on accuracy on unseen compositions. 
This confirms the importance of restricting the search space under a criterion taking into account the probability of a composition of being a distractor, such as our feasibility scores.

\begin{figure*}[t]
 \includegraphics[width=\textwidth]{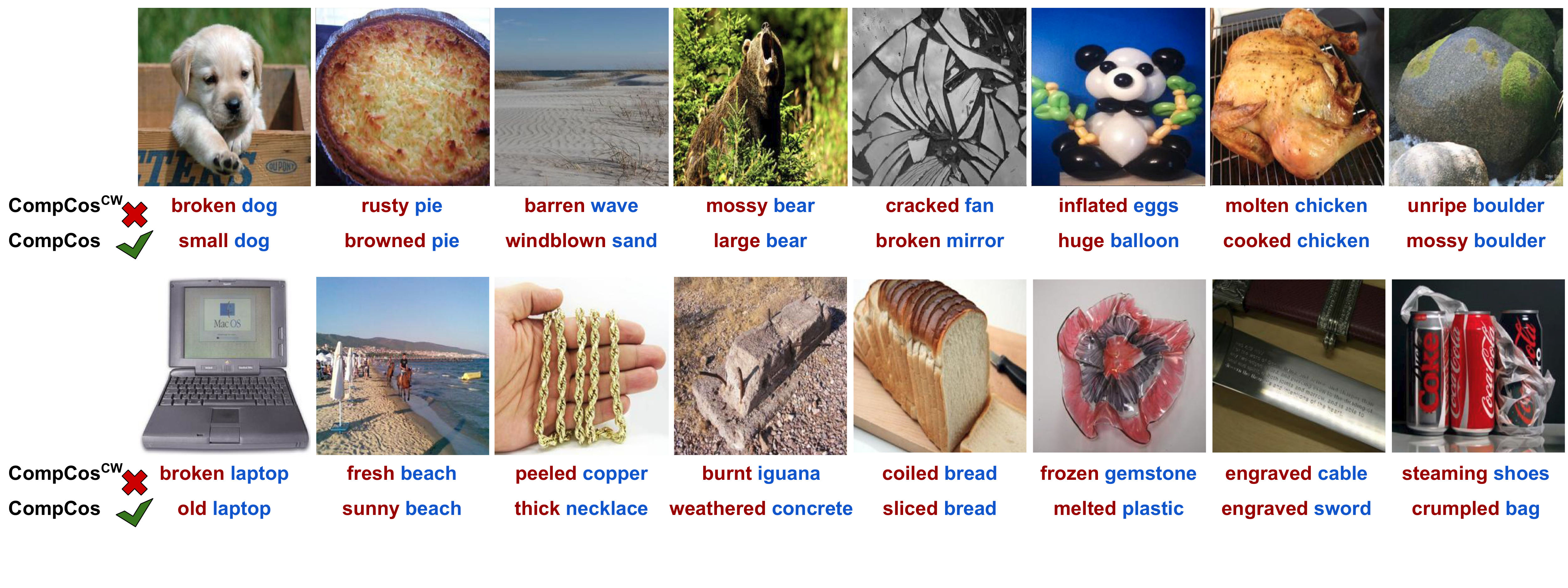}\myvspace{-25pt}
\caption{Examples correct predictions of \ours\ in the OW-CZSL scenario when the \oursClosed\ fails. The first row shows the predictions of the closed world model, the bottom row shows the results of \ours. The images are randomly selected.} 
\myvspace{-10pt}

\label{fig:qualitative}
\end{figure*}

An interesting observation is that applying our feasibility-based binary masks on top of other approaches (\ie LE+, TMN, and SymNet) is beneficial. SymNet, modeling states and objects separately, sees a minor increase in performance in HM while maintaining the AUC and unseen accuracy. However, LE+ and TMN, which learn a joint compatibility between compositions and images, see a big increase in performance with the introduction of feasibility. TMN improves from 0.1 to 0.4 in the AUC while seeing a big increase in the HM from 1.7 to 4.1. Similarly, LE+ improves from 0.3 AUC to 0.5 with a significant increase in the HM, from 3.2 to 4.6.

\subsection{Qualitative results} 
We show example composition predictions for a set of images of our \oursClosed and our \ours\ in the open world setting. Furthermore, we show examples of most- and least-feasible compositions determined by our model.

\myparagraph{Compositions Corrected due to Feasibility Scores.} 
We qualitatively analyze the reasons for the improvements of \ours\ over \oursClosed, by looking at the example predictions of both models on simple sample images of MIT states. We compare predictions on samples where the closed world model is ``distracted'' by a distractor while the open world model is able to predict the correct class label.  

As shown in Figure \ref{fig:qualitative}, the closed world model is generally not capable of dealing with the presence of distractors. For instance, there are cases where the object prediction is correct (\eg \textit{broken dog}, \textit{molten chicken}, \textit{unripe boulder}, \textit{mossy bear}, \textit{rusty pie}) but the associated state is not only wrong but also making the compositions unfeasible. In other cases, the state prediction is almost correct (\eg \textit{cracked fan},\textit{barren wave}, \textit{inflated eggs}) but the associated object is unrelated, making the composition unfeasible. All these problems are less severe in our full \ours\ model since our feasibility-driven objective helps in isolating the implausible distractors in the shared embedding space, reducing the possibility to predict them. 

\setlength{\tabcolsep}{2pt}
\renewcommand{\arraystretch}{1.2}
\begin{table}[t]
    \centering
    \resizebox{\linewidth}{!}{
    \begin{tabular}{ l  l  l }
    \hline
        & \multicolumn{2}{c}{\textbf{Compositions}} \\
       & Most Feasible (Top-1)  &  Least Feasible (Bottom-1)   \\\hline
       & browned tomato& short lead  \\
        &caramelized potato & cloudy gemstone \\
         & thawed meat &standing vegetable \\
          &small dog &  full nut \\
           &large animal&  blunt milk  \\
           \hline
        \textbf{Objects} & \multicolumn{2}{c}{\textbf{States}} \\
        & Most Feasible (Top-3)  &  Least Feasible (Bottom-3) \\
        \hline
         tomato& browned, peeled, diced & tight, full, standing\\
         dog &small, old, young &fallen, toppled, standing\\
         cat & wrinkled, huge, large  &viscous, smooth, runny\\
         laptop&small, shattered, modern & cloudy, sunny, dull\\
         camera&tiny, huge, broken&diced, caramelized, cloudy\\
    \end{tabular}
    }
    \vspace{1pt}
    \caption{Unseen compositions wrt their feasibility scores (Top: Top-5 compositions on the left, Least-5 on the right; Bottom: Top-3 highest and Bottom-3 lowest feasible state per object.}
    \myvspace{-12pt}
    \label{tab:most-feasible-general}
\end{table}

\myparagraph{Discovered Most and Least Feasible Compositions.} The most crucial advantage of our method is its ability to estimate the feasibility of each unseen composition, to later inject these estimates into the learning process. Our assumption is that our procedure described in Section \ref{sec:compcos-open} is robust enough to model which compositions should be more feasible in the {compositional} space and which should not, isolating the latter in the shared embedding space. We would like to highlight that here we are focusing mainly on visual information to extract the relationships. This information can in principle be coupled with knowledge bases (\ie \cite{liu2004conceptnet}) and language models (\ie \cite{wang2019language}) to further refine the scores.

Table \ref{tab:most-feasible-general} (Top) shows qualitative examples of the most and least feasible compositions discovered by \ours. As an example, it correctly ranks \textit{browned tomato} and \textit{small dog} as one of the most feasible compositions, while \textit{full nut} and \textit{blunt milk} among the least feasible ones. Since the dataset has a lot of food classes, we see that the top and bottom are mostly populated by them. However, the presence of relevant and irrelevant states with these objects is promising and shows the potential of our feasibility estimation strategy.

Table \ref{tab:most-feasible-general} (Bottom) shows the top-3 most feasible compositions and bottom-3 least feasible compositions given five randomly selected objects. These objects specific results show a tendency of the model to relate feasibility scores to the subgroups of classes. For instance, cooking states are considered as unfeasible for standard objects (\eg \textit{diced camera}) as well as atmospheric conditions (\eg \textit{sunny laptop}). Similarly, states usually associated with substances are considered unfeasible for animals (\eg \textit{runny cat}). At the same time, size and ages are mostly {linked with} animals (\eg \textit{young dog}) while cooking states are correctly associated with food (\eg \textit{diced tomato}). Interestingly, the top states for \textit{cat} are all present with \textit{dog} as seen compositions, thus the similarities between the two classes has been used to transfer these states from \textit{dog} to \textit{cat}, following Eq.~\eqref{eq:obj}. 
 
\section{Conclusions}
In this work, we propose a new benchmark for CZSL that extends the problem from the closed world to an open world where all the combinations of states and objects could potentially exist. We show that state-of-the-art methods fall short in this setting as the number of unseen compositions significantly increases. We argue that not all combinations are valid classes but it is unrealistic to assume that test set pairs are the only valid compositions. We propose a way to model the feasibility of a state-object composition by using the visual information available in the training set. This feasibility is independent of an external knowledge base and can be directly incorporated in the optimization process.
We propose a novel model, \ours, that incorporates this feasibility and achieves state-of-the-art performance in both closed and open world on two real-world datasets. 

\vspace{-5pt}
\myparagraph{\newline Acknowledgments}
This work has been partially funded by the ERC (853489 - DEXIM) and by the DFG (2064/1 – Project number 390727645).

{\small
\bibliographystyle{ieee_fullname}
\bibliography{egbib}
}

\section*{\Large Appendix}
\appendix

\section{Expanded Results}
\subsection{Comparison with the State of the Art}

\setlength{\tabcolsep}{2pt}
\renewcommand{\arraystretch}{1.2}
\begin{table*}[t]
    \centering
    \resizebox{\linewidth}{!}{
    \begin{tabular}{l| c c c c c c | c c c c c c | c c c c c c | c c c c c c}
    \multirow{3}{*}{\textbf{Method}} 
    & \multicolumn{12}{c}{\textbf{Closed World}}& \multicolumn{12}{|c}{\textbf{Open World}}\\
    & \multicolumn{6}{c}{\textbf{MIT states}}& \multicolumn{6}{c}{\textbf{UT Zappos}}& \multicolumn{6}{|c}{\textbf{MIT states}}& \multicolumn{6}{c}{\textbf{UT Zappos}}\\
                                                &Sta. &Obj.    &S   & U    & HM    & auc &Sta. &Obj.    &S   & U    & HM    & auc &Sta. &Obj.    &S   & U    & HM    & auc &Sta. &Obj.    &S   & U    & HM    & auc \\\hline
     AoP\cite{nagarajan2018attributeasoperators}&21.1   & 23.6      &14.3   &17.4       &9.9    &1.6 &38.9   & 69.9      &\textbf{{59.8}}   &54.2       &40.8    &25.9
     &15.4   & 20.0      &16.6   &5.7       &4.7    &0.7 & 25.7  &   61.3    &  50.9 &  34.2     & 29.4   & 13.7\\
     LE+\cite{misra2017redwine}                 &23.5   & 26.3      &15.0   &20.1       &10.7   &2.0 &41.2   & 69.3      &53.0   &{61.9}       &41.0   &25.7&
     10.9   & 21.5      &14.2   &2.5       &2.7   &0.3  & 
     \textbf{38.1}  &  68.2     & \textbf{60.4}  &   36.5    & 30.5  & 16.3            \\
     TMN\cite{purushwalkam2019tmn}              &23.3   & 26.5      &20.2   &20.1       &13.0   &2.9 &40.8   & 69.5      &58.7   &60.0       &\textbf{45.0 }  &\textbf{29.3}
     &  6.1   &   15.9   &  12.6        &  0.9     &   1.2        &   0.1  &  14.6    &         61.5 &     55.9  &18.1           & 21.7       &  8.4\\
     SymNet\cite{li2020symnet}                  &26.3   & 28.3      &24.2   &\textbf{25.2}       &{16.1}   &3.0 &41.3   & {68.6}      &49.8   &{57.4}       &{40.4}   &23.4
     &17.0   & 26.3      &21.4   &7.0       &5.8   &0.8              &33.2   &70.0       &53.3   &44.6       &34.5   &18.5\\\hline
     \textbf{\oursClosed}                             &\textbf{27.9}   & \textbf{31.8}      &{25.3}   &24.6       &\textbf{16.4}   &\textbf{4.5} &\textbf{{44.7}}   & \textbf{73.5}      &\textbf{59.8}   &\textbf{62.5}       &{43.1}   &{28.7}
     &13.9  & \textbf{28.2}      &25.3   &5.5       &5.9   &0.9 &33.8   &72.4       &59.8   &45.6   &36.3   &20.8\\
          \textbf{\ours}                             &{26.7}   & {30.0}      &\textbf{25.6}   &22.7       &{15.6}   &{4.1} &{{44.0}}   & {73.3}      &{59.3}   &{61.5}       &{40.7}   &{27.1}
     &\textbf{18.8}   & {27.7}      &\textbf{25.4}   &\textbf{10.0}       &\textbf{8.9}   &\textbf{1.6} & 35.1 & \textbf{72.4} & 59.3 & \textbf{46.8} & \textbf{36.9} & \textbf{21.3}\\
    \end{tabular}}
    \vspace{1pt}
    \caption{Closed and Open World CZSL results on MIT states and UT Zappos. We measure states (Sta.) and objects (Obj.) accuracy on the primitives, best seen (S) and unseen accuracy (U), best harmonic mean (HM), and area under the curve (auc) on the compositions.}
    \label{tab:sota-expanded}
\end{table*}

In Table 1 of the main paper, we reported the comparison between \ours  and the state of the art, in both closed and open world settings. As highlighted in the methodological section, closed and open world are different problems with different challenges (\ie bias on the seen classes for the first, presence of distractors in the second). For this reason, in the closed world experiments, we reported the results of the closed world version of our model (Section 3.2), while our full model is used for the more complex OW-CZSL (Section 3.3). Here, we expand the table, reporting the results of the closed world (\oursClosed) and full (\ours) versions of our model for both closed and open world scenarios.

Table \ref{tab:sota-expanded} shows the complete results for both MIT states and UT Zappos. As we can see, both versions of our model achieve competitive results on the closed world scenario and in both datasets. In this setting, our full model, \ours, achieves slightly lower performance than \oursClosed, with a 4.1 AUC vs the 4.5 AUC of our closed world counterpart on MIT states, and a 27.1 vs 28.7 of AUC on UT Zappos. This is because our full model focuses less on balancing seen and unseen compositions (the crucial aspect of standard closed world CZSL) but mostly on the margin between feasible and unfeasible compositions. This latter goal is not helpful in the closed world setting, where the subset of feasible compositions seen at test time is known a priori.  Nevertheless, the performance of \ours\ largely surpasses the previous state of the art on MIT states in AUC, with a 1.1 increase of AUC over SymNet.

On the other hand, if we exclude \ours, our closed world model (\oursClosed) achieves the highest AUC when applied in the open world scenario, in both datasets. In particular, it is comparable to SymNet on MIT states (0.8 vs 0.9 AUC) while surpassing it by 2.3 AUC on UT Zappos. On MIT states, it achieves a lower performance unseen accuracy with respect to SymNet (\ie 5.5\% vs 7.0\%). We believe this is because SymNet is already robust to the inclusion of distractors by modeling objects and states separately during inference. Nevertheless, our full approach is the best in all compositional metrics and in both datasets. In particular, on MIT states it improves \oursClosed\ by 4.5\% on best unseen accuracy, 3.0\% on best harmonic mean, and 0.8 of AUC. This confirms the importance of including the feasibility of each composition during training.

\subsection{Ablating Masked Inference}
In the main paper (Table 3), we tested the impact of thresholding the feasibility scores to explicitly exclude unfeasible compositions from the output space of the model (Section 3.3, Eq.~(6)). In particular, Table 3 shows how the binary masks obtained from \ours\ can greatly improve the performance of our closed world model, \oursClosed, and other approaches (\ie LabelEmbed+, TMN) while being only slightly beneficial to more robust ones such as our full method \ours\ and SymNet. 

Here we analyze whether the effect of the mask is linked to limiting the output space of the model or to their ability to excluding the majority of the distractors (\ie less feasible compositions). To test this, we apply to the output space of \ours\ and \oursClosed, two additional binary masks. The first is obtained by thresholding the feasibility scores using their median (\textit{median}), keeping as valid unseen compositions all the ones with the score above the median. The second is the reverse, \ie we keep as valid all the seen compositions, and all the unseen compositions whose feasibility scores are below the median (\ie \textit{inv. median}). 

What we expect is that, if the feasibility scores are not meaningful, distractors are equally excluded, no matter if we consider the top half or the bottom half of the scores. If this happens, the performance boost would be only linked to the fact that we exclude a portion of the output space, and not to the actual unfeasibility of the excluded compositions. Consequently, we would expect \ours\ and \oursClosed\ to achieve the same results when either \textit{median} or \textit{inv. median} are applied as masks on the output space. 

The results of this analysis are reported in Table \ref{tab:ablation-inference-median}, where we report also the results of not excluding any composition (\textit{-}) and of the best threshold value (\textit{best}). As the Table shows, the performance gaps are very large if we take as valid the compositions having the top or the bottom half of the scores. In particular, in \oursClosed\ performance go from 1.3 to 0.03 in AUC, 7.5\% to 0.3\% in harmonic mean, and from 6.9\% to 0.1\% in best unseen accuracy. \ours\ shows a similar behavior, with the AUC going from 2.2 to 0.06, the harmonic mean from 10.9\% to 0.6\%, and the best unseen accuracy from 11.1\% to 0.4\%. These results clearly demonstrate that i) the boost brought by masking the output space is linked to the exclusion of unfeasible compositions rather than a simple reduction of the search space; ii) feasibility scores are meaningful, with the feasible compositions tending to receive the top-50\% of the feasibility scores.

\begin{figure}
  {\includegraphics[width=\columnwidth,height=.33\columnwidth]{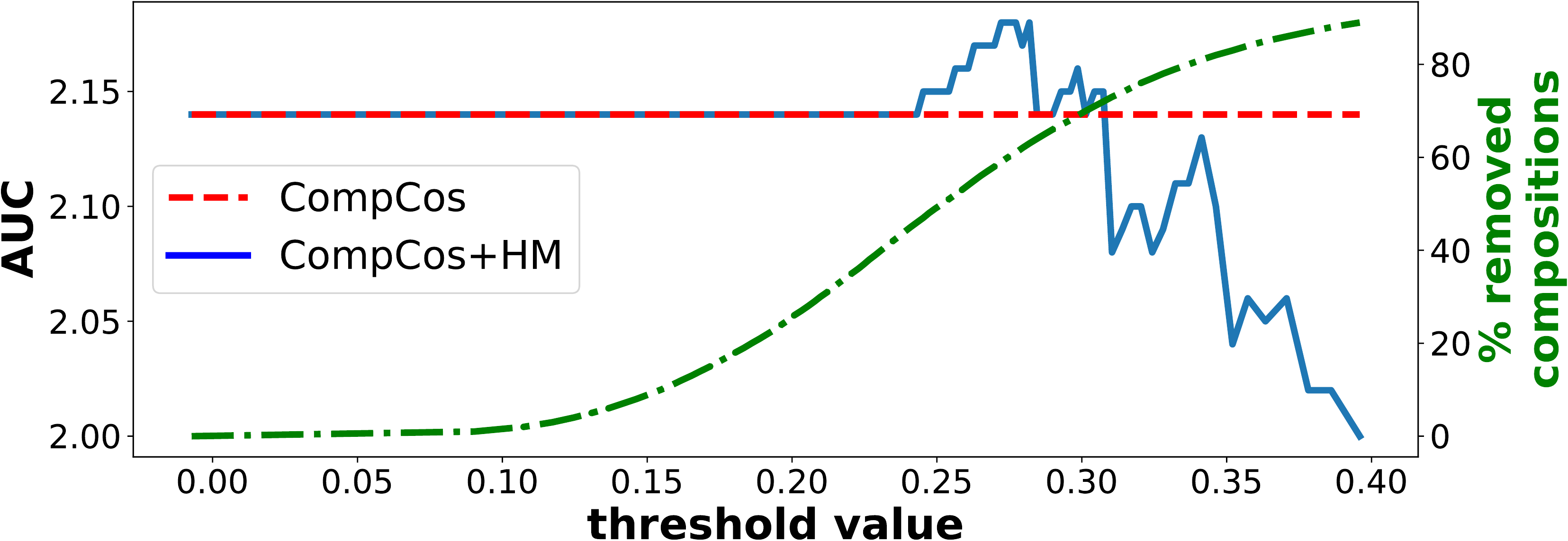}}
    \caption{
           \ours: AUC vs hard masking threshold on MIT states' validation set. The green line denotes the percentage of removed compositions at a given threshold value. 
    } 
    \label{fig:threshold}
\end{figure}

Finally, in Figure \ref{fig:threshold} we analyze the impact of that the hard masking threshold on \ours on the validation set of MIT states. As the figure shows, low threshold values remove a few percentage (green) of the compositions and the AUC is comparable to the base model with no hard masking (red). By increasing the threshold, the AUC increases up to the point where the output space is overly restricted and also (feasible) compositions of the dataset 
are discarded. Indeed, hard masking can work only if 
the similarity scores (and the ranking of compositions they produce) are meaningful, otherwise even low values would mask out feasible compositions from
the dataset, harming the model's performance.

\setlength{\tabcolsep}{3pt}
\renewcommand{\arraystretch}{1.2}
\begin{table}[t]
    \centering
    \begin{tabular}{l| c | c c c c}
       & Mask   &Seen   & Unseen    & HM    & AUC\\\hline
                  
          \multirow{2}{*}{\oursClosed}  & - &  \multirow{4}{*}{28.0} &6.0 &7.0 &1.2\\
          & median  &  &6.9&7.5& 1.3 \\
          & inv. median &  &0.1&0.3& .03 \\
          & best  &  &8.1&8.7& 1.6 \\
          \hline
                  
         \multirow{2}{*}{\ours}  &-&\multirow{4}{*}{27.1}&{11.0}&10.8& {2.1}\\
          & median  &  &11.1&10.9& 2.2 \\
          & inv. median &  &0.4&0.6& .06 \\
          &  best & &11.2&11.0&2.2  \\
    \end{tabular}
    \vspace{1pt}
    \caption{Results on MIT states validation set for applying our feasibility-based binary masks ($f_{\text{HARD}}$) on \oursClosed\ and \ours\ with different strategies.}
    \label{tab:ablation-inference-median}
\end{table}

\section{Additional Qualitative Results}

\subsection{Feasibility Scores}
In this section, we focus on MIT states and we report additional qualitative analyses on the most and least feasible compositions, as for the feasibility scores computed by our model. In particular, in Table \ref{tab:most-feasible-obj} we show the top-3 and bottom-3 states associated to 25 randomly selected objects, while in Table \ref{tab:most-feasible-att} we show the top-3 and bottom-3 objects associated to 25 randomly selected states. 

Similarly to the analysis of the main paper, in Table \ref{tab:most-feasible-obj} we can see how the highest feasibility scores are generally linked to related sub-categories of objects/states. For instance, \textit{gate} is related to conservation-oriented status (\ie \textit{cracked}, \textit{dented}) while cooking states (\ie \textit{cooked}, \textit{raw}, \textit{diced}) are considered its most unfeasible. A similar observation applies to \textit{necklace}, associated to conservation status (\ie \textit{pierced}, \textit{scratched}) while states related to atmospheric conditions (\ie \textit{cloudy}, \textit{open} related to \textit{sky}) are considered unfeasible. Cooking states are the most feasible for \textit{chicken} (\ie \textit{diced}, \textit{thawed}, \textit{cooked}) while cloth states are related to \textit{jacket}  (\ie \textit{crumpled}, \textit{wrinkled}, \textit{torn}), as expected.

In Table \ref{tab:most-feasible-att}, we show a different analysis, \ie we check what are the most/least feasible objects given a state. Even in this case, we see a similar trend, with food (\ie \textit{potato}, \textit{tomato}, \textit{sauce}) associated as feasible to food-related states (\eg \textit{cooked}, \textit{mashed}, \textit{moldy}, \textit{unripe}) while clothing items (\eg \textit{shirt}, \textit{jacket}, \textit{dress}) associated as feasible to clothing-related states (\ie \textit{draped}, \textit{loose}, \textit{ripped}). On the other hand, we can see how environments (\eg \textit{ocean}, \textit{beach}) are associated as feasible to their meteorological state (\eg \textit{sunny}, \textit{cloudy}) but not to manipulation ones (\eg \textit{bent}, \textit{pressed}). 

Overall, Tables \ref{tab:most-feasible-obj} and \ref{tab:most-feasible-att} show how the feasibility scores capture the subgroups to which objects/states belong.  This suggests that when the feasibility scores are introduced as margins within the model, related subgroups are enforced to be closer in the output space than unrelated ones,  
improving the discrimination capabilities of the model.

\setlength{\tabcolsep}{2pt}
\renewcommand{\arraystretch}{1.18}
\begin{table}[t]
    \centering
    \resizebox{\linewidth}{!}{
    \begin{tabular}{ l  l  l }
    \hline

        \textbf{Objects} & \multicolumn{2}{c}{\textbf{States}} \\
        & Most Feasible (Top-3)  &  Least Feasible (Bottom-3) \\\hline
aluminum & unpainted, thin, coiled & full, closed, young\\
apple & peeled, caramelized, diced & full, standing, short\\
bathroom & grimy, cluttered, steaming & fallen, unripe, cooked\\
beef & browned, sliced, steaming & standing, cluttered, fallen\\
blade & broken, straight, shiny & young, sunny, cooked\\
bronze & melted, crushed, pressed & full, open, winding\\
cave & tiny, verdant, damp & blunt, whipped, diced\\
chicken & diced, thawed, cooked & standing, open, closed\\
dress & wrinkled, ripped, folded & cloudy, open, closed\\
fence & crinkled, weathered, thick & short, full, cooked\\
garlic & browned, sliced, squished & standing, full, closed\\
gate & closed, cracked, dented & cooked, raw, diced\\
glasses & broken, dented, crushed & cloudy, smooth, sunny\\
island & small, foggy, huge & blunt, short, open\\
jacket & crumpled, wrinkled, torn & cloudy, young, full\\
library & huge, modern, heavy & blunt, cooked, viscous\\
necklace & thick, pierced, scratched & cloudy, full, open\\
potato & caramelized, sliced, mashed & full, short, standing\\
ribbon & creased, frayed, thick & cloudy, sunny, cooked\\
rope & thick, curved, frayed & modern, ripe, cluttered\\
shower & dirty, empty, tiny & standing, unripe, young\\
steps & small, large, dented & blunt, fresh, raw\\
stream & foggy, verdant, dry & young, standing, closed\\
sword & shattered, blunt, rusty & ripe, full, cooked\\
wool & thick, crumpled, ruffled & full, fallen, closed\\
        \hline
    \end{tabular}
    }
    \vspace{1pt}
    \caption{Unseen compositions wrt their feasibility scores: Top-3 highest and Bottom-3 lowest feasible state per object.}
    \label{tab:most-feasible-obj}
\end{table}

\setlength{\tabcolsep}{2pt}
\renewcommand{\arraystretch}{1.2}
\begin{table}[t]
    \centering
    \resizebox{\linewidth}{!}{
    \begin{tabular}{ l  l  l }
    \hline
        \textbf{States} & \multicolumn{2}{c}{\textbf{Objects}} \\
        & Most Feasible (Top-3)  &  Least Feasible (Bottom-3) \\
        \hline
        ancient & town, house, road & sauce, well, foam\\
barren & canyon, river, jungle & penny, handle, paint\\
bright & coast, cloud, island & handle, key, drum\\
closed & gate, window, garage & persimmon, berry, copper\\
cloudy & coast, shore, beach & gemstone, penny, shoes\\
cooked & meat, salmon, chicken & field, gate, library\\
creased & shirt, newspaper, shorts & animal, fire, lightning\\
fresh & vegetable, pasta, meat & handle, key, steps\\
loose & shorts, jacket, clothes & butter, seafood, salmon\\
mashed & tomato, potato, fruit & book, stream, deck\\
moldy & apple, pear, sauce & handle, drum, key\\
molten & candy, butter, milk & shoes, cat, animal\\
painted & wood, granite, metal & fig, well, book\\
peeled & tomato, apple, pear & cat, cave, ocean\\
pressed & steel, cotton, silk & field, well, beach\\
ripped & jacket, hat, dress & cat, seafood, cave\\
shiny & blade, stone, sword & city, well, animal\\
squished & vegetable, garlic, bean & road, shore, cat\\
sunny & beach, ocean, sea & card, penny, wire\\
thawed & meat, seafood, chicken & wave, handle, cat\\
unpainted & aluminum, roof, metal & animal, book, lightning\\
unripe & persimmon, vegetable, potato & gear, shoes, phone\\
verdant & valley, pond, coast & penny, keyboard, book\\
winding & highway, tube, wire & bear, armor, beef\\
worn & pants, clothes, shorts & seafood, animal, fire\\
\hline
    \end{tabular}
    }
    \vspace{1pt}
    \caption{Unseen compositions wrt their feasibility scores: Top-3 highest and Bottom-3 lowest feasible object per state.}
    \label{tab:most-feasible-att}
\end{table}

\subsection{Qualitative examples}

\begin{figure*}[t]
 \includegraphics[width=\textwidth]{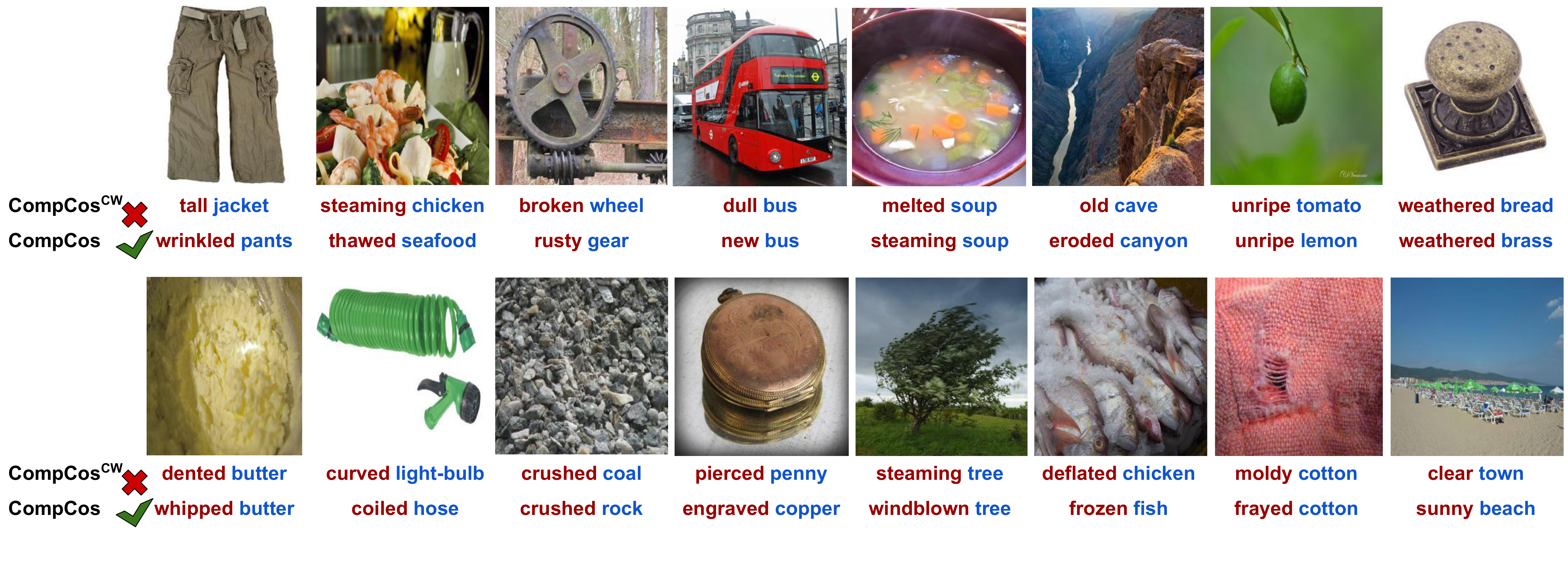}\vspace{-18pt}
\caption{Examples correct predictions of \ours\ in the OW-CZSL scenario when the \oursClosed\ fails. The first row shows the predictions of the closed world model, the bottom row shows the results of \ours. The images are randomly selected.} 
\label{fig:qualitative2}
\end{figure*}

In this subsection, we report additional qualitative examples, comparing the predictions of our full model \ours\ with its closed world counterpart, \oursClosed, on MIT states. Similarly to Figure 3 of the main paper, in Figure \ref{fig:qualitative2}, we show examples of images misclassified by \oursClosed\ but correctly classified by \ours. The figure confirms that \oursClosed\ is less capable than \ours\ to deal with the presence of distractors.
In fact, there are cases where \oursClosed\ either misclassifies the object (\eg \textit{cave} vs \textit{canyon}, \textit{bread} vs \textit{brass}), the state (\eg \textit{steaming} vs \textit{thawed}, \textit{moldy} vs \textit{frayed}) or both terms of the composition (\ie \textit{broken well} vs \textit{rusty gear}, \textit{curved light-bulb} vs \textit{coiled hose}). While in some cases the answer is close to the correct one (\eg \textit{unripe tomato} vs \textit{unripe lemon}, \textit{crushed coal} vs \textit{crushed rock}) in others the error is mainly caused by the presence of less feasible compositions in the output space (\eg \textit{deflated chicken}, \textit{melted soup}). These compositions are not correctly isolated by \oursClosed, thus they hamper the discriminative capability of the model itself. This does not happen with our full model \ours\, where unfeasible compositions are better modeled and isolated in the compositional space.

\begin{figure*}[t]
 \includegraphics[width=\textwidth]{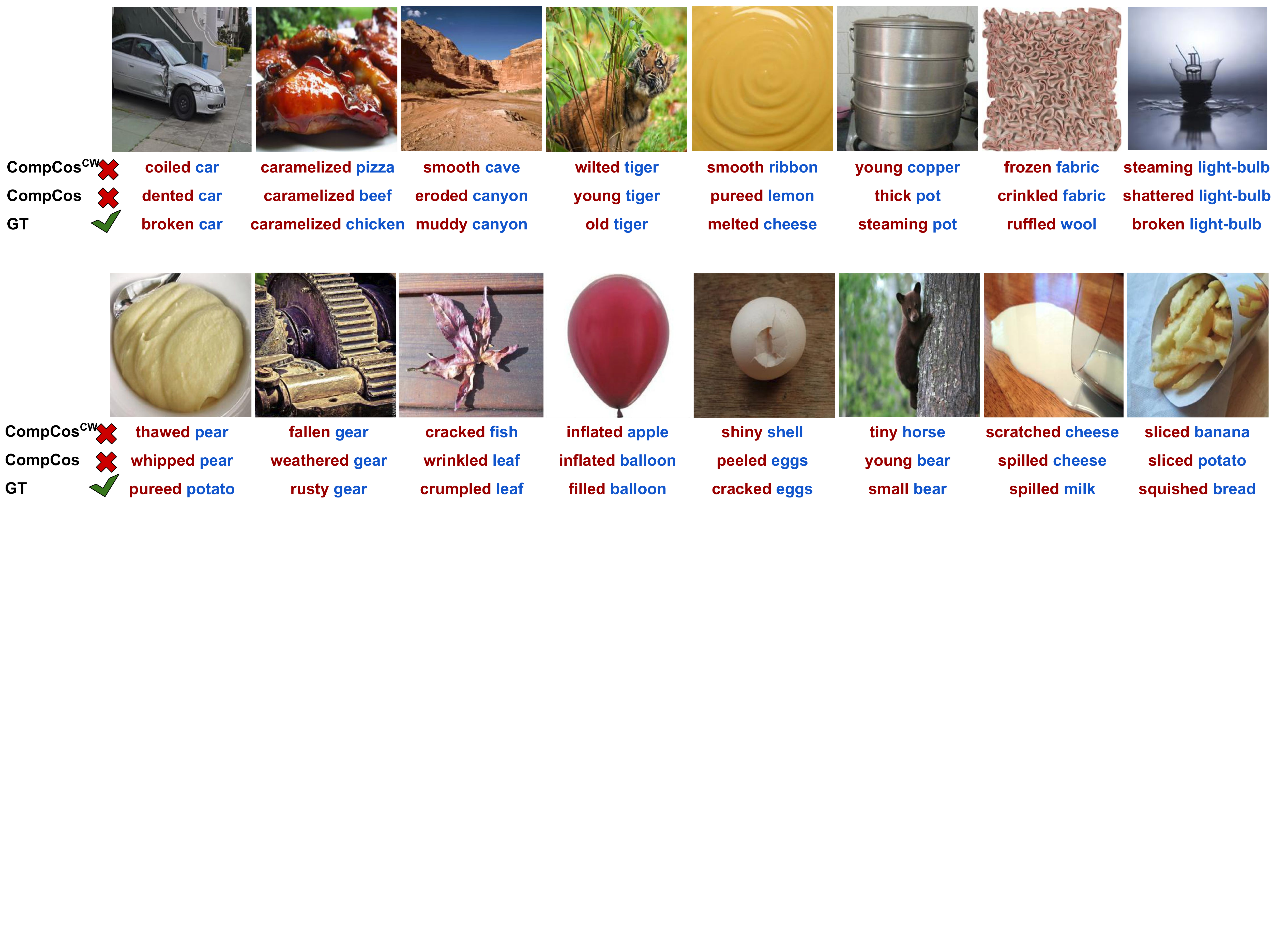}\vspace{-18pt}
\vspace{-160pt}
\caption{Examples of wrong predictions of \ours\ and \oursClosed\ in the OW-CZSL scenario. The first row shows the predictions of the closed world model, the second row shows the results of \ours, the third row the ground-truth (GT). Images are randomly selected.} 
\label{fig:qualitative-wrong}
\end{figure*}

As a second analysis, in Figure \ref{fig:qualitative-wrong}, we show some examples where both \ours\ and \oursClosed\ are incorrect. Even in this case, it is possible to highlight the differences among the answers given by the two models. \oursClosed\, being less capable of dealing with the presence of distractors, tends to give implausible answers in some cases (\eg \textit{inflated apple}, \textit{coiled car}, \textit{young copper}, \textit{wilted tiger}). On the other hand, our full model still gives plausible answers, despite those being different from the ground-truth. For instance, while \oursClosed\ misclassifies the \textit{caramelized chicken} as \textit{caramelized pizza}, \ours\ classifies it as \textit{caramelized beef}, with the actual object (\ie \textit{beef} vs \textit{chicken}) being hardly distinguishable from the picture, even for a human. There are other examples in which \ours\ recognizes a state close to the one of the ground-truth (\eg \textit{inflated} vs \textit{filled}, \textit{shattered} vs \textit{broken}, \textit{weathered} vs \textit{rusty}, \textit{eroded} vs \textit{muddy}) or a plausible composition given the content of the image \textit{\eg \textit{crinkled fabric}, \textit{spilled cheese}}. We also reported one example where the prediction of our model is correct while the annotation being incorrect (\ie \textit{sliced potato} vs \textit{squished bread}) and some where the prediction of the model is compatible with the content of the image, as well as the ground-truth (\eg \textit{young bear}, \textit{dented car}, \textit{thick pot}). We found the last observations to be particularly interesting, highlighting another problem that future works should tackle in CZSL: the presence of multiple states in a single image.

\end{document}